\documentclass[letterpaper, 10 pt, conference]{ieeeconf}  
\pdfoutput=1

\IEEEoverridecommandlockouts                              

\usepackage{amsmath} 
\usepackage{amssymb}  
\usepackage{algorithm, algorithmicx, algpseudocode}

\usepackage{subcaption}
\usepackage[pdftex]{graphicx}
\usepackage{import}
\usepackage[dvipsnames]{xcolor}
\usepackage{placeins}
\usepackage[font=small, labelfont=bf]{caption}
\usepackage{cite}

\newcommand{\rtext}[1]{ {\color{red}#1} }
\newcommand{\gtext}[1]{ {\color{OliveGreen}#1} }
\newcommand{\btext}[1]{ {\color{blue}#1} }

\def\Vec#1{\!\!\hbox{$#1$\kern-0.38em\lower0.85em\hbox{$\vec{}\,$}}\,}%
\newcommand{\bbm}{\begin{bmatrix}}
\newcommand{\ebm}{\end{bmatrix}}

\newcommand{\x}{\mathbf{x}}

\newcommand{\mb}[1]{\mathbf{#1}}

\title{\LARGE \bf
Experience Recommendation for Long Term Safe Learning-based Model Predictive Control in Changing Operating Conditions
}

\author{Christopher D. McKinnon and Angela P.~Schoellig
\thanks{The authors are with the Dynamic Systems Lab ({www.dynsyslab.org}) at the University of Toronto Institute for Aerospace Studies (UTIAS), Canada. Email: {chris.mckinnon@mail.utoronto.ca, schoellig@utias.utoronto.ca}}%
\thanks{This work was supported in part by the Natural Sciences and Engineering Research Council of Canada under the grant RGPIN-2014-04634 and by the Connaught New Researcher Award.}}

\begin{document}

\maketitle
\thispagestyle{empty}
\pagestyle{empty}

\begin{abstract}

Learning has propelled the cutting edge of performance in robotic control to new heights, allowing robots to operate with high performance in conditions that were previously unimaginable. The majority of the work, however, assumes that the unknown parts are static or slowly changing. This limits them to static or slowly changing environments. However, in the real world, a robot may experience various unknown conditions.  This paper presents a method to extend an existing single mode GP-based safe learning controller to learn an increasing number of non-linear models for the robot dynamics. We show that this approach enables a robot to re-use past experience from a large number of previously visited operating conditions, and to safely adapt when a new and distinct operating condition is encountered. This allows the robot to achieve safety and high performance in an large number of operating conditions that do not have to be specified ahead of time. Our approach runs independently from the controller, imposing no additional computation time on the control loop regardless of the number of previous operating conditions considered. We demonstrate the effectiveness of our approach in experiment on a 900\,kg ground robot with both physical and artificial changes to its dynamics. All of our experiments are conducted using vision for localization.
 \end{abstract}

\section{INTRODUCTION}

At the core of most control algorithms in robotics is a model that captures the relationship between the state, the input, and the dynamics of a robotic system. The model can be used to optimize a reward function and to ensure that the system achieves its goals in a safe and reliable way \cite{kober2013reinforcementSurvey, Berkenkamp2015SafeRobust}. If the model for the system is partially unknown, the reward function can incorporate an element to encourage exploration of the system dynamics \cite{AbeelOptimism2015, AbeelModelBasedOptamism15}. This establishes a better mapping between the state, input, and dynamics, such that the controller can later exploit well-known, high-reward actions \cite{AbeelOptimism2015}. An accurate assessment of the risk associated with taking a control action, especially if it has not been taken before, is important during the exploration process \cite{Berkenkamp2015SafeControllerIROS,Ostafew2015ICRA}. Using an assessment of this risk to ensure safety is known as safe learning.

\begin{figure}
	\centering
	\includegraphics[width=0.5\textwidth]{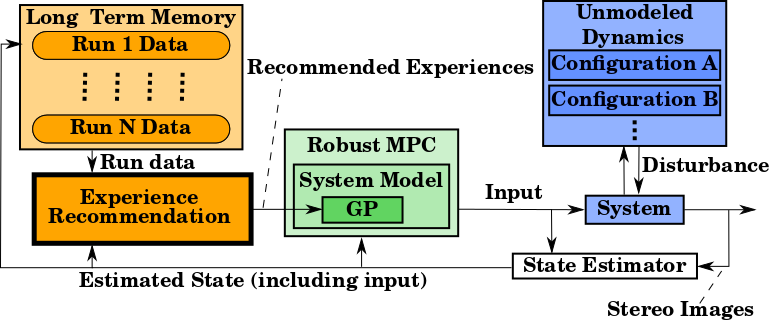}
	\vspace{-10pt}
	\caption{Block diagram showing the proposed experience recommendation in closed-loop with a safe controller. The system dynamics can change from one run to another. The proposed experience recommendation chooses the experiences from previous runs that best match the current dynamics. These experiences are used in a Gaussian Process (GP) to model the system dynamics. The robust Model Predictive Controller (MPC) is the controller from \cite{Ostafew2015RCLBMPC}.}
	\label{fig:BlockDiagram}
\end{figure}

Safe learning methods generally incorporate an approximate initial guess for the system dynamics with some bounds on the modelling error incurred in the approximation \cite{Ostafew2015RCLBMPC, Tomlin2013RobustLearningMPC}. A learning term then refines the initial guess over time using experience data to better approximate the true dynamics. The goal is to guarantee that the system does not violate safety constraints (e.g., limits on the control input or path tracking error) while achieving the control objective (e.g., following a path) and at the same time improving the model of system and, consequently, its task performance over time. Most learning algorithms learn a single model for system dynamics or use multiple models that are trained ahead of time based on appropriate training data from operating the robot in all relevant conditions \cite{Ostafew2015RCLBMPC, AbeelModelBasedOptamism15, InfoTheoreticRL2017, Luders2013robustParafoil, Li2017Impromptu, Roy2013DynamicSafePlanning}. This presents a challenge for robots that are deployed into a wide range of operating conditions which may not all be known ahead of time. 


This paper builds on work in \cite{Ostafew2015RCLBMPC}, which proposes a robust, learning-based Model Predicitve Controller (MPC) for repetitive path following using a vision-based localization algorithm \cite{paton2017expanding}. The controller developed in \cite{Ostafew2015RCLBMPC} uses local Gaussian Process (GP) regression to construct a model for the dynamics at each time-step. These models are valid over a small section of the path around the vehicle's current position. They are based on a fixed number of training points, or experiences, so have fixed computational cost. Our contribution is to generalize this approach for long-term safe learning with large and small, repeated changes in the dynamics that depend on the operating conditions or physical configuration of the robot (see Fig.~\ref{fig:BlockDiagram}). Examples are weather, terrain conditions \cite{Angelova2008Thesis}, or payload configuration \cite{McKinnon2017MMLearning}. The proposed method builds local GPs of  the robot dynamics on each previous run to select experiences from the run with the most similar dynamics. These experiences are used to construct the GP used for control. This allows the GP used for control to leverage knowledge related to changes in dynamics that are both sudden and gradual as long as a similar change has been observed in the past. We use the same GP hyper-parameters as the GP in the controller. Accordingly, we introduce no additional model parameters and are able to assess how likely it is that the GP constructed from previous experiences will satisfy the assumptions of the safe controller, namely that the 3$\sigma$ bounds on uncertainty are an accurate upper bound on model error. The local GPs used by our method are inexpensive to compute enabling the robot to learn over and leverage data from a large number of runs without having to `forget' previous experiences. This is a significant advantage over previous methods.

\section{RELATED WORK}

Learning control has received a great amount of attention in recent years, most notably in the case of single-mode learning control. This is the broad class of learning methods that assumes the true (but initially unknown) mapping between the state at time $k$, $\x_k$, and the input $\mb{u}_k$, and the next state, $\x_{k+1}$, is one-to-one, or at least normally distributed according to some underlying process, $\x_{k+1}\sim \mathcal{N}(f(\x_k,\mb{u}_k),\boldsymbol\sigma)$. Recent developments have contributed safety guarantees \cite{Tomlin2013RobustLearningMPC} and demonstrated impressive results in improved path following \cite{Ostafew2014ICRA}. 

Multimodal safe control and path planning has also received a growing amount of attention. Applications include safely gliding a parafoil under a variety of wind conditions~\cite{Luders2013robustParafoil} and planning safe paths among uncertain agents such as pedestrians or automobiles \cite{Roy2013DynamicSafePlanning}. The assumption and challenge in these cases is that the environment or obstacles in the environment have hidden states that change the dynamics cannot be measured directly. Similar to our method, the algorithms try to infer this hidden state based on available observations and use this for safe planning. An additional feature of our approach is that we attempt to build this model online and do not explicitly require discrete changes in dynamics.

Recent results in single-mode, safe learning control have taken great steps to improve performance while maintaining bounds on modelling error and therefore safety. Approaches by  \cite{Ostafew2014ICRA, Ostafew2015RCLBMPC, Tomlin2013reducing, Tomlin2012QuadExperiment, Abeel2014CableControlCorr} use GPs as corrective terms for approximate prior models and update them over time as more experience is gathered. 
A GP assumes a single underlying function with additive Gaussian noise. They are a perfect tool when there is only one mode for the dynamics or the mode can be measured directly. Bounds on the model error are used to allocate margin on safety constraints such that the system is robust to this model error. It is essential for safety that the bounds from the GP,
usually some multiple of the standard deviation, bound the true model error. It is essential for high performance that the bounds are not unnecessarily conservative. A GP assumes a single underlying function with additive Gaussian noise, and is excellent when there is only one mode for the dynamics; however, if there are multiple dynamic modes (e.g., caused by driving both on snow and asphalt), the single GP must have overly conservative bounds to account for the dynamics in all modes (which may be significantly different), and may learn some combination of the dynamics in each mode which is sub-optimal in either mode. This is of particular concern
to algorithms that update the GP online.

One model that exhibits especially good real-time performance and has been demonstrated in several real-world examples is presented in \cite{Ostafew2015RCLBMPC}. This approach continually reconstructs the GP disturbance model based on a fixed number of data points, to ensure the process model can be evaluated in constant time even if new experience is added. Storing the data in first-in-first-out bins of fixed size allows the algorithm to update the data used in the GP in real time. If the mode changes, the model un-learns the existing mode by over-writing all of that data and re-learns the new mode. During this process, it suffers from the same problems related to hyper-parameters as mentioned above including either requiring over-conservative bounds to accommodate multiple modes, or have bounds that are realistic for a single-mode, but are unsafe while the model transitions between modes and is using data from more than one mode. Our method aims to overcome these limitations by only choosing data that is relevant to the current mode.

In addition to the single-mode, safe learning controllers, multimodal algorithms exist which identify a number of dynamic modes ahead of time using labelled or unlabelled training data and switch to the most likely model during operation \cite{jo2012IMMEKF, Luders2013robustParafoil, Roy2013DynamicSafePlanning, calandra2015learning, Mouret2017PriorSelection}. This allows them to maintain persistent knowledge of a robot's dynamics across a wide range of operating conditions.  Inferring the correct mode from measurements during operation allows them to maintain a high level of performance and robustness even when the mode is not directly observed. The method proposed in~\cite{Fox20XXDPHMM} for linear systems even infers the number of modes at training time. These approaches do, however, require that the number of modes and/or training data from each mode be available ahead of time, which can be a challenging task in robotics. In contrast, our method does not require the number of modes or training data from each mode to be available ahead of time. Rather, it learns new dynamics as they arise during operation.  

In our previous work, we presented an approach for learning multi-modal dynamics by combining GPs and the Dirichlet Process, which is used in Bayesian non-parametric clustering models \cite{McKinnon2017MMLearning}. This allowed the robot to learn a new GP model for novel operating conditions and leverage an existing GP when the robot re-visited an operating condition. The GPs used in \cite{McKinnon2017MMLearning} represented the dynamics over the entire region of the state space and therefore required a large number of training points to be effective. These GPs could therefore not be used directly in the controller, which is limited to GPs with only a small number of training points. The proposed method overcomes this by keeping the size of the GP used for control constant. In addition, the previous approach only used a relative measure of model quality. That is, it would only switch to the prior, safe mode if that described the current dynamics better than the existing set of GP experts. In this work, we add an explicit check to ensure that the assumptions made by the safe controller are likely to be valid.


In light of the current approaches and their limitations, the goal of this paper is to present a method for adapting to multiple dynamic modes over a long period of time with guarantees on safety using a realistic and computationally efficient representation of the system dynamics (including predictive uncertainty). The aim is to design an algorithm for life long model learning to achieve excellence in the relevant operating conditions regardless of whether they are known ahead of time.


\section{PROBLEM STATEMENT}

The goal of this work is to learn a model for the dynamics of a ground robot performing a repetitive, path-following task. The robot may be subjected to large changes in its dynamics due to factors such as payload, terrain, weather, or tyre pressure changes. We assume that these factors cannot be measured directly. The algorithm should scale to long-term operation and take advantage of repeated runs in the same operating conditions. The model should also include a reasonable estimate of model uncertainty that acts as an upper bound on model error at all times.

Further assumptions can be summarized as follows:
\begin{itemize}
  \item The mapping $(\mathbf{u}_k,\x_k) \to \x_{k+1}$  can be modelled as a GP for a single run.
  \item The operating condition is constant over a short time horizon.
  \item The number of operating conditions and the mapping $(\mathbf{u}_k,\x_k) \to \x_{k+1}$ for each is not known ahead of time.
\end{itemize}

A short time horizon could be similar to the horizon considered for MPC.

The system can be modelled by some nominal dynamics $\mathbf{f}(\x_k,\mb{u}_k)$ with additive, initially unknown  dynamics $ \mb{g}^c(\mb{a}_k)$ that are specific to discrete or continuous operating conditions $c$ and depend on features $\mb{a}_k$, so
\begin{equation}
  \mb{\x}_{k+1} = \mb{f}(\mb{x}_k,\mb{u}_k) + \mb{g}^c(\mb{a}_k).
  \label{eqn:MotionModel}
\end{equation}

The unknown dynamics are assumed to be a deterministic function with additive, zero-mean, Gaussian noise,
\begin{equation}
    \mb{g}^c(\mb{a}_k) =  \mb{g}^c_0(\mb{a}_k) + \boldsymbol\eta^c,
    \label{eqn:UnmodeledDynamics}
\end{equation}
where $\boldsymbol\eta^c \sim \mathcal{N}(0,\boldsymbol\Sigma_{\eta}^c)$, and $ \boldsymbol\Sigma_{\eta}^c$ is the measurement noise covariance.



\section{METHODOLOGY}

In this section, we present our approach for long-term, safe learning control. Our approach makes extensive use of local GPs to model the robot dynamics.




\subsection{Gaussian Process (GP) Disturbance Model}
We model the unknown dynamics, $\mb{g}(\cdot)$, as a GP based on past observations. We drop the $(\cdot)^c$ for notational convenience because we learn a GP for each operating condition separately. Since there are many good references on GPs \cite{GPforML2006}, here we provide only a high-level sketch. The learned model depends on previously gathered experiences which are assembled from measurements of the state denoted by $\hat{\x}$ and the input $\mb{u}$ using \eqref{eqn:MotionModel}, so that
\begin{align}
    \hat{\mb{g}}(\mb{a}_{k-1}) = \hat{\x}_k - \mb{f}(\hat{\x}_{k-1}, \mb{u}_{k-1}).
    \label{eqn:GeneralExperienceFormula}
\end{align}
The resulting pair, $\left\lbrace\mb{a}_{k-1},\hat{\mb{g}}(\mb{a}_{k-1})\right\rbrace$, forms an individual experience. For simplicity, we model each dimension of the disturbance using a separate GP. Below, we derive the equations for a single dimension of $\mb{g}(\cdot)$ denoted by $g(\cdot)$.

A GP is a distribution over functions given past experiences, $\mathcal{D} =\{\mb{a}_i, \hat{g}(\mb{a}_i)\}_{i=1}^m$, and kernel hyper-parameters. 

We assume the experiences are noisy observations of the true function $g(\mb{a}_k)$; this is, $\hat{g}(\mb{a}_k) = g(\mb{a}_k) + \eta_{\eta}$ where $\eta_{\eta} \sim \mathcal{N}(0, \sigma_{\eta}^2)$.  The posterior distribution is characterized by a mean and variance which can be queried at any point $\mb{a}_*$ using
\begin{align}
  \label{eqn:GPmean}
  \mu(\mb{a}_*) &=  \mb{k}(\mb{a}_*)\mb{K}^{-1}\hat{\mb{g}},\\
  \sigma^2(\mb{a}_*) &= \kappa(\mb{a}_*, \mb{a}_*) - \mb{k}(\mb{a}_*)\mb{K}^{-1}\mb{k}(\mb{a}_*)^T,
  \label{eqn:GPvariance}
\end{align}
where $\hat{\mb{g}} = [\hat{g}(\mb{a}_1),...,\hat{g}(\mb{a}_m)]^T$ is the vector of observed function values, the covariance matrix $\mb{K}\in\mathbb{R}^{m\times m}$ has entries $[\mb{K}(\mb{a}_i,\mb{a}_j)] = \kappa(\mb{a}_i,\mb{a}_j) + \sigma_{\eta}^2 \delta_{ij}$, where~$\delta_{ij}$ is the Kronecker delta, and the vector $\mb{k}(\mb{a}_*)=[\kappa(\mb{a}_*, \mb{a}_1), ..., \kappa(\mb{a}_*, \mb{a}_m)]$ contains the covariances between the new test point $\mb{a}_*$ and the observed data points $\mathcal{D}$. For this work, we use the squared exponential kernel,
\begin{align}
    \kappa(\mb{a}_i, \mb{a}_j) =\sigma_f^2 \exp\left( - \frac{1}{2 }(\mb{a}_i - \mb{a}_j)^T \mb{L}^{-2}(\mb{a}_i - \mb{a}_j)\right)
    \label{eqn:Kernel}
\end{align}
because of its success in modelling robot dynamics \cite{Ostafew2014ICRA, Ostafew2015ICRA,Ostafew2015RCLBMPC, Tomlin2012QuadExperiment}. The hyper-parameters are the diagonal matrix, $\mb{L}$, of length-scales which are inversely related to the importance of each element of $\mb{a}$, and the process noise variance, $\sigma_f^2$, which is the variance of the prior family of functions represented by $g(\cdot)$.

As training data is added to a particular GP, uncertainty is reduced and the posterior distribution of the GP specializes to a particular family of functions which represents the system dynamics in a particular operating condition. The job of the experience recommendation, which is the contribution of this work, is to choose training data that results in the GP specializing to the functions that represent the robot dynamics in the current operating condition.

\subsection{Controller and System Model}

For this work, we use the robust model predictive controller from \cite{Ostafew2015RCLBMPC} and refer the reader to this paper for details. 

\subsection{Data Management}

\begin{figure}
  \centering
  \includegraphics[width=0.45\textwidth]{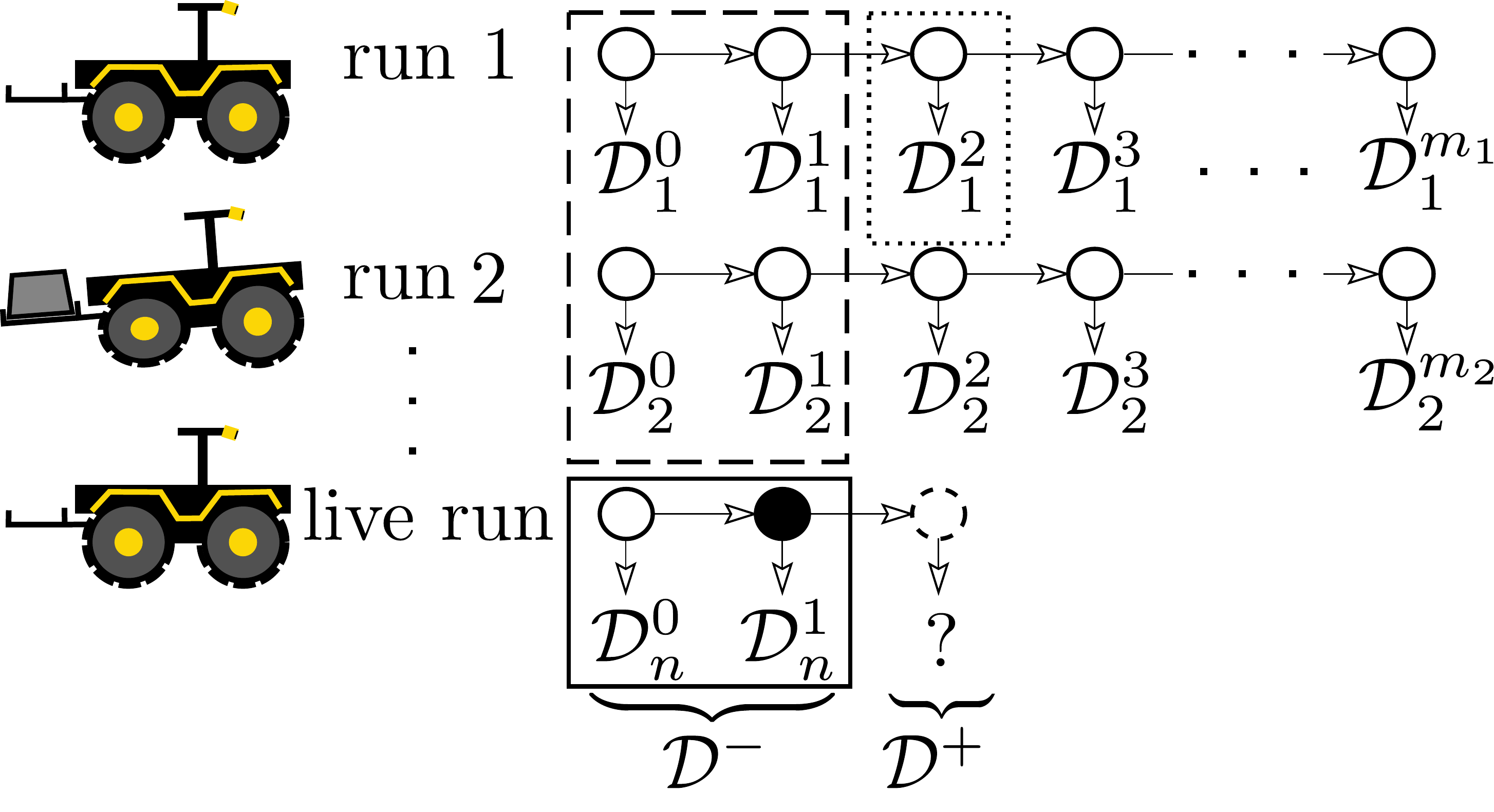}
  \caption{Runs 1 and 2 represent previous autonomous traverses of the path and the live run represents the current traverse. The filled black circle indicates the current position of the vehicle and the dotted circle indicates the predicted position of the vehicle at the end of the MPC prediction horizon. Data, $\mathcal{D}_r^v$, for each run, $r$, is stored in small sets at each vertex, $v$, represented by a circle. The goal is to use recent data from the live run (solid box) to assess the similarity of the current dynamics to the dynamics in all previous runs along the same section of the path (dashed box). This is used to recommend experiences from a run with similar dynamics (dotted box) and construct a predictive model for the dynamics on the upcoming section of the path.}
  \label{fig:ExpRecConceptualDrawing}
\end{figure}

The purpose of our method is to construct the best possible model of the system dynamics for MPC. MPC uses the dynamics over the upcoming section of the path to compute the control. We use data from the recently traversed part of the path to determine which past runs are relevant and can be used for the MPC prediction model. Referring to Fig.~\ref{fig:ExpRecConceptualDrawing}, for each previous run, $i$, we use data, $\mathcal{D}_i^-$, from the recently traversed section of the path to construct a local GP, $\hat{g}_i^-$. Each of these GPs is then used to generate predictions for the mean and variance, $\left\lbrace \hat{\mu}_{i,j}^-, \hat{\sigma}_{i,j}^-\right\rbrace_{j=1..m_n}$, at each of the $m_n$ $\mb{a}_{n,j}$'s in recent experience from the current run, $\mathcal{D}_n^-$. These estimates are used to identify the most similar run to the current run and reject runs that are substantially different from the current run.

To update the control GP (see Fig. \ref{fig:BlockDiagram}) with new points to model the dynamics on the upcoming section of the path, we randomly draw ten experiences (if available) from the most similar run in a window ahead of the vehicle overlapping with the MPC prediction horizon. These experiences are added to the set of experiences already in the control GP. Fifty experiences (if available) are then randomly chosen from this combined set to get a new set of experiences for the control GP \cite{McKinnon2017MMLearning}. If no runs are recommended, we randomly remove ten data points from the set in the GP used for control. If this happens several times in a row, the control GP will quickly revert to the prior which acts as a `safe mode' since its uncertainty bounds are the most conservative.  We only use experiences from the most similar run, however, we could easily make use of experiences from multiple runs if not enough experience was available from one run.

The control GP uses fifty points to allow the controller to run at 10 Hz with a 1.5 second look-ahead. Updating the GP incrementally helps the model change smoothly which in turn results in smoother control actions. We found to be helpful during our experiments.

\subsection{Run Rejection Criterion}
\label{sec:RunRejection}

Our first step is to eliminate runs where the dynamics are so different from the current dynamics that using data from these runs is likely to result in model errors that violate assumptions made by the safe controller. In particular, we must ensure that the $3\sigma$ bounds on the prediction from the GP are a reasonable upper bound on the model error \cite{Ostafew2015RCLBMPC}.

To do this for candidate run $i$, we test whether the proportion of samples from $\mathcal{D}_n^-$ that lie further than  $3\hat{\sigma}_{i,j}^-$  from the corresponding predicted $\hat{\mu}_{i,j}^-$ is significantly higher than would be expected by chance. We do this using the binomial test.

Let $m_n$ be the number of input-output pairs in $\mathcal{D}_n^-$ and $N_{\textnormal{out}}$ be the number of outliers, points outside of the predicted $3\sigma$ bounds, according to the predictions from $\hat{g}_i^-$. The binomial distribution $B(x,m_n, p)$ describes the probability of drawing exactly $x$ outliers from $m_n$ independent samples  where the probability of drawing an outlier is $p$. We calculate the probability of $N_{\textnormal{out}}$ or more outliers using
\begin{equation}
	p(N_{\textnormal{out}}\text{ or more}) = \sum_{x=N_{\textnormal{out}}}^{m_n}B(x,m_n,p).
\end{equation}
We reject the run if $p(N_{\textnormal{out}}\text{ or more})  < \alpha$ where $\alpha$ is the significance level, or the probability of falsely rejecting a run. We chose a $5\%$ significance level for our experiments. This may be reduced to avoid falsely rejecting runs, or increased to be more conservative. Since $3\hat{\sigma}_{i,j}^-$ is such a conservative bound, changing $\alpha$ only changes the allowed number of outliers by one or two for $m_n=100$ so the algorithm is not very sensitive to this parameter.

Any runs that make it past this step are considered as candidates for drawing experiences to model the dynamics over the upcoming section of the path.


\subsection{Run Similarity Measure}
\label{sec:RunSimilarity}

Our next step is to identify which runs are most similar to the current run given recent data from the live run, $\mathcal{D}_n^-$ and corresponding predictions $\{\hat{\mu}^-_{i,j}, \hat{\sigma}^-_{i,j}\}_{j=1..m_n}$ from the local GP for each candidate run, $i$.

We assume that the vehicle can be in a different operating condition for each run, and that one set of GP hyper-parameters is sufficient to describe the robot dynamics in each operating condition when operating conditions are considered separately. We can then compute the posterior probability that the current dynamics are from the same operating condition, $c$, as run $i$ using
\begin{equation}
 p(c=i|\mathcal{D}_n^-, \hat{g}_i^-) \propto p(\mathcal{D}_n^- | \hat{g}_i^-) p(c=i).
\end{equation}
The second term on the right is the prior, which we assume to be equal for all runs. The first term on the right is the probability of recent experiences if the operating conditions are the same as run $i$. Assuming each experience is independent, this is 
\begin{equation}
	 p(\mathcal{D}_n^- | \hat{g}_i^-) = \prod_{j=1}^{m_n} p(\hat{g}(\mb{a}_{n,j})|\hat{\mu}^-_{i,j}, \hat{\sigma}^-_{i,j}),
\end{equation}
where $\hat{\mu}^-_{i,j}$ and $\hat{\sigma}^-_{i,j}$ are the predicted mean and variance of the GP for run $i$ evaluated at point $\mb{a}_{n,j}$, which is in $\mathcal{D}_n^-$. Similar to our previous work \cite{McKinnon2017MMLearning}, we reject any run that has lower probability than the GP prior of generating $\mathcal{D}_n^-$. This is to ensure that experience added to the GP for control is likely to improve the performance beyond what could be achieved with no experience at all.

In our implementation, we use the log-probability to avoid numerical issues and do not normalize the posterior since we are only interested in finding the most likely run and not the actual probability distribution. We denote the un-normalized log-probability for run $i$ with $L_i$. The run with the largest $L_i$ is chosen as the recommended run.

\subsection{Overview of the Algorithm}

Putting the components above together, we arrive at the experience recommendation algorithm, Alg. \ref{alg:ExperienceRec}.

\begin{algorithm}
\caption{Overview of the experience recommendation algorithm.}
\label{alg:ExperienceRec} 
\begin{algorithmic}
  \State $n \gets$ live run number
  \State $\alpha \gets$ significance level for binomial test
  \State $\mathcal{D}^- \Leftarrow \left\lbrace \mathcal{D}_i^-, i=1..n-1 \right\rbrace$ \Comment{data from candidate runs}
  \State $\hat{g}^-_i \gets$ Fit a GP to each candidate run using data in $\mathcal{D}^- $
  \State $scored\_runs = \left\lbrace \right\rbrace$
  \State $\mathcal{A}_n^-, \mathcal{G}_n^- \gets$ GP inputs from $\mathcal{D}_n^-$, GP outputs from $\mathcal{D}_n^-$
  \For{$i=1..n-1$} 
    \State $\left\lbrace \hat{\mu}_{i,j}^-, \hat{\sigma}_{i,j}^-\right\rbrace_{j=1..m_i} \gets \hat{g}_i^-(\mathcal{A}_n^-)$
    \State $N_{\textnormal{out}} \gets \#$ outliers given $\left\lbrace \hat{\mu}_{i,j}^-, \hat{\sigma}_{i,j}^-\right\rbrace_{j=1..m_n} , \mathcal{G}_n^-$
  	\State $p_b \gets$ Binomial test probability given $N_{\textnormal{out}}$
    \If{$p_b < \alpha$}
    	\State \textbf{continue}
    \EndIf
    \State Compute $L_i$ given $\left\lbrace \hat{\mu}_{i,j}^-, \hat{\sigma}_{i,j}^-\right\rbrace_{j=1..m_n} , \mathcal{G}_n^-$
    \State Append ($i$, $L_i$) to $scored\_runs$
  \EndFor
  \State $\hat{\mathcal{D}}_n^+ \gets$ experiences in the control GP
  \If{$scored\_runs$ is empty}
      \State Remove 10 points from $\hat{\mathcal{D}}_n^+$
  \Else
  	 \State $i^* \gets$ run with largest $L_i$
  	 \State Randomly add 10 experiences from run $i^*$ to $\hat{\mathcal{D}}_n^+$
  	 \State Randomly remove 10 points from $\hat{\mathcal{D}}_n^+$
  \EndIf
   \State Construct a new GP for control using $\hat{\mathcal{D}}_n^+$
\end{algorithmic}
\end{algorithm}



\section{Experiments}
\label{sec:ExperimentalResults}

\begin{figure}
	\centering
	\includegraphics[width=0.5\textwidth]{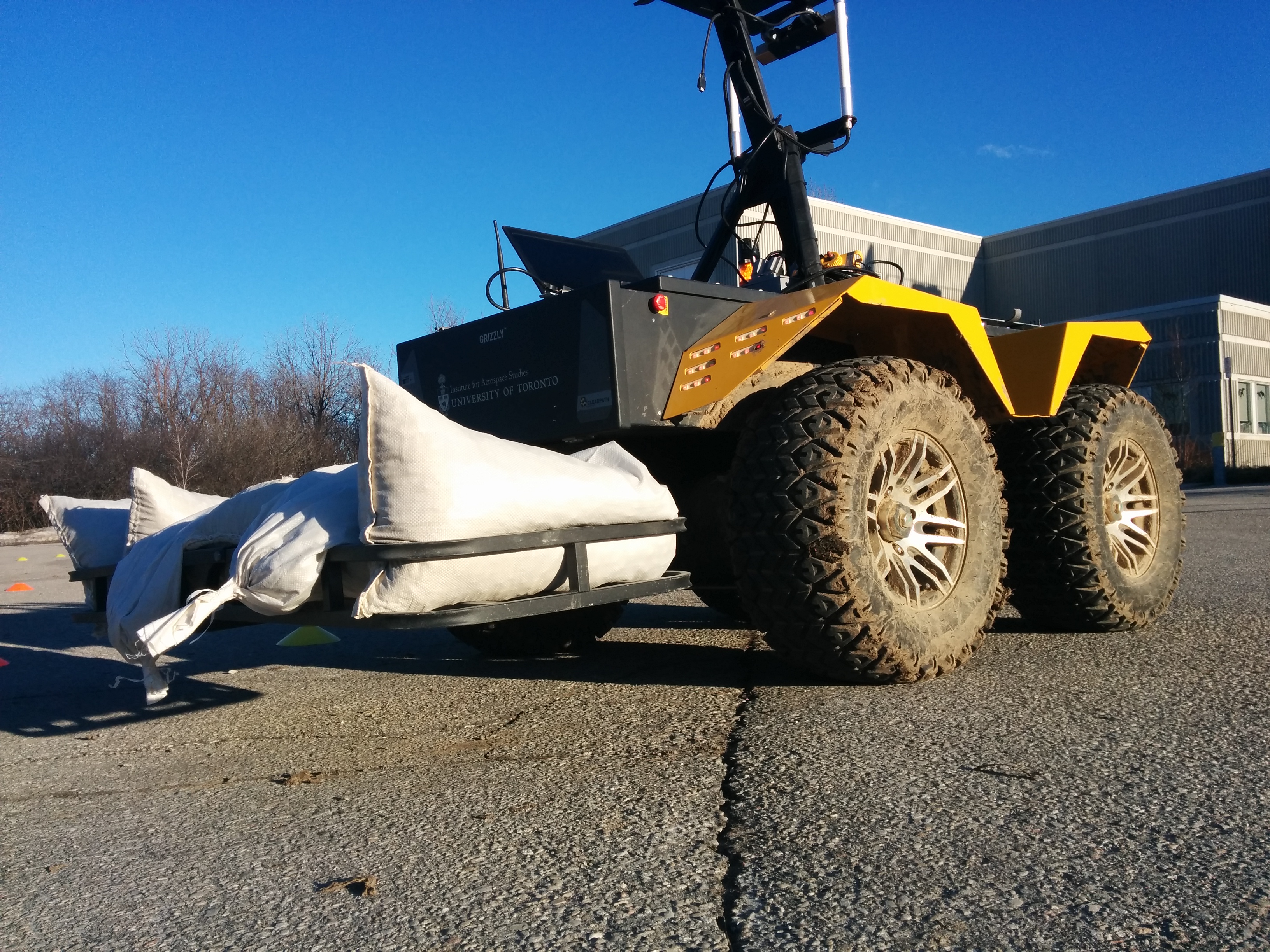}
	\caption{Clearpath Grizzly  in the \emph{loaded} configuration with six bags of gravel in the cargo carrier. Each bag weighs approximately 30\,kg for a total payload of 180\,kg.}
	\label{fig:LoadedGrizzly}
\end{figure}

Experiments were conducted on a 900\,kg Clearpath Grizzly skid-steer ground robot shown in Fig.\,\ref{fig:LoadedGrizzly}. We tested our algorithm with the Grizzly in three configurations. First, the \textit{nominal} configuration, with no changes to the vehicle. Second, the \textit{loaded} configuration, with six bags of gravel, weighing approximately 30\,kg each, in a cargo carrier mounted on the Grizzly (see Fig. \ref{fig:LoadedGrizzly}). Finally, the \textit{altered} configuration, where the rotational rate commands were multiplied by 0.7. Compared to the \textit{nominal} configuration, the \textit{loaded} configuration results in over-steer and the \textit{altered} configuration results in under-steer.

Our first experiment was conducted in a parking lot on a 42\,m long course. During this experiment, the configuration was switched between the \textit{nominal} and \textit{altered} configurations. We compare our proposed method to a baseline method, which uses only experiences from the most recent run. We conducted three runs in each configuration to allow the baseline method  to converge to the dynamics in each configuration before switching. This serves as a simple case to demonstrate a few features of our algorithm.

Our second experiment was also conducted in a parking lot on a similar course. However, we switched between all three configurations after only two runs in each over a total of 30 runs to compare our method to the baseline method during long term operations. This was to demonstrate that the proposed method continues to learn during long term operation and maintains high performance in all three configurations regardless of the order of configuration switches.

\subsection{Implementation}
Our algorithm was implemented in C++ and can process up to 300 runs in a single thread at 2 Hz on an Intel i7 2.70 GHz 8 core processor with 16 GB of RAM. This number is extrapolated based on the fact that the computational cost of the proposed method scales linearly with the number of runs. We consider the last three seconds of data (30 samples) from the live run for $\mathcal{D}_n^-$. The experience recommendation process runs in a separate thread to the controller. Therefore, it does not add any computation time to the control loop and can run at a different rate to process more runs. Our controller relies on a vision-based system, Visual Teach and Repeat \cite{paton2017expanding}, for localization.

\subsection{System Model and Controller Parameters}
Our process model is the unicycle with an additive GP learning term \cite{Ostafew2015RCLBMPC},
\begin{equation}
	\bbm x_{k+1} \\ y_{k+1} \\ \theta_{k+1} \ebm = \underbrace{\bbm x_{k} \\ y_{k} \\ \theta_{k} \ebm +  dt\bbm\cos{\theta_k} & 0 \\ \sin\theta_k & 0 \\ 0 & 1 \ebm \bbm v^{cmd}_k \\ \omega^{cmd}_k \ebm}_{\text{unicycle}} + dt \underbrace{\bbm g_x(\mb{a}_k) \\ g_y(\mb{a}_k) \\ g_{\theta}(\mb{a}_k) \ebm}_{\text{learning term}},
\end{equation}
where $v_k^{cmd}$ and $\omega_k^{cmd}$ are the commanded speed and turn rate, $x$ and $y$ are the position in the reference frame, $\theta_k$ is the orientation, and $dt$ is the time-step of the controller.
The cost function for MPC is a quadratic penalty on lateral error, heading error, $\omega_k^{cmd}$, $(v_k^{cmd}- v_k^d)$ where $v_k^d$ is the desired speed, $\dot{\omega}_k^{cmd}$, and $\dot{v}_k^{cmd}$. The respective weights are 500, 35, 5, 4, 1000 and 500.  The desired speed was set at 1.5 m/s for all of our experiments.

\subsection{Model Predictive Performance}

In order to evaluate the quality of the models constructed using our method, we first compared the multi-step prediction performance of a GP constructed using our experience recommendation method to a GP constructed using experiences from the most recent run.  We consider the rotational dynamics, because they differ the most between configurations.

To measure the accuracy of the prediction of the mean, we use the Multi-step RMS Error (M-RMSE) between the prediction made over the look-ahead horizon in MPC and the measured state at these times. To measure the accuracy of the error bound, we will use the Multi-step RMS Z-score (RMSZ) of the prediction at the future time-steps.

The M-RMSZ for a prediction of $p$ time-steps is
\begin{align}
    M-RMSZ_k &= \sqrt{\frac{1}{p}\sum_{j=k+1}^{k+p}\frac{(\dot{\theta}^{true}_j - \mu_{\dot{\theta}}(\mb{a}_j))^2}{\sigma_{\dot{\theta}}(\mb{a}_j)^2}}
\end{align}
where $\mu_{\dot{\theta}}(\mb{a}_j)$ and $\sigma_{\dot{\theta}}(\mb{a}_j)$ are the mean and standard deviation of the predicted rotational rate evaluated at the predicted GP input $\mb{a}_j$. The true measurements of angular velocity, $\dot{\theta}_j^{true}$, are used for comparison. An M-RMSZ around one is ideal. A larger M-RMSZ around two indicates that the model is over-confident and M-RMSZ less than one indicates that the model is conservative.

\begin{figure}
\centering
\scriptsize
\def\svgwidth{0.475\textwidth}
\graphicspath{{figs/Feb24_results/}}
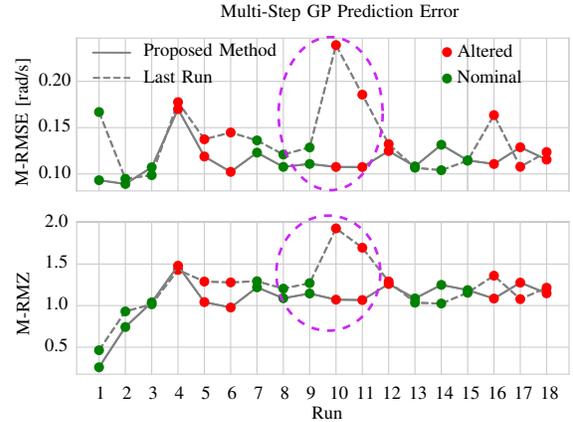\caption{This figure shows how the proposed method  improves the GP multi-step prediction performance compared to using experiences from the last run, especially during runs 8-11, indicated by the purple dotted circle. The configuration is switched between the \gtext{\textbf{\textit{nominal}}} configuration (green) and the \rtext{\textbf{\textit{altered}}} configuration (red). The lower M-RMSZ and higher M-RMSE on run the third transition during run 16 for the Last Run method indicates that the controller exploited actions where the model was more uncertain for this run.}
\vspace{-0.25cm}
\label{fig:PerformanceComparison}
\end{figure}

Figure \ref{fig:PerformanceComparison} shows that the M-RMSE after a transition is up to 2.5 times higher when using experiences from the last run compared to the proposed method. The M-RMSZ shows a similar trend with the proposed method continually closer to one than the baseline method. We ignore run one for this comparison because the robot did not have any experience.

\subsection{Closed Loop Performance}
\label{sec:ClosedLoopPerformance, Experiment 1}

To assess the impact of our method on closed loop performance, we compare the speed, control cost, and tracking error using our method compared to when the GP is constructed using experiences from the last run. 

\begin{figure}
\centering
\scriptsize
\def\svgwidth{0.475\textwidth}
\graphicspath{{figs/Feb24_results/}}
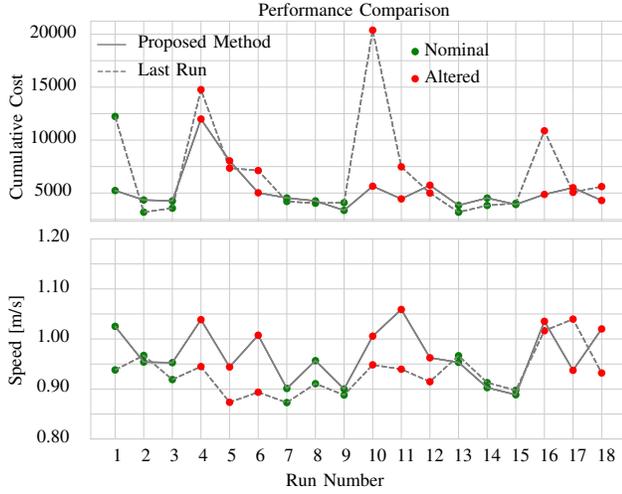  \caption{The cumulative step cost calculated using the MPC penalty function on the states and actions the system actually visited during a run, and the average speed during each run. Runs in the \gtext{\textbf{\textit{nominal}}} configuration are green and runs in the \rtext{\textbf{\textit{altered}}} configuration are red. The proposed method reduces the control cost compared to using experiences from the last run. The average speed when using the proposed method is slightly higher because the controller does not have to slow down to make large corrections as often.}
  \label{fig:LearningTransitionsSpeedCost}
\end{figure}

Figure \ref{fig:LearningTransitionsSpeedCost} shows that after transitions from the nominal configuration to the altered configuration, the proposed method significantly lowers the control cost compared to the baseline method. This cost comes from large lateral tracking errors due to under-steering. We do not observe the same difference when transitioning to the nominal configuration because the vehicle will tend to over-steer in this case, and the penalty function in MPC penalizes the magnitude of the turn rate commands, which naturally prevents the vehicle from over-steering. Choosing a different control cost may result in an increased cost for transitioning the configuration either way.

\subsection{Experience Recommendation by Configuration}
\begin{figure}
\centering
\scriptsize
\def\svgwidth{0.475\textwidth}
\graphicspath{{figs/Feb24_results/}}
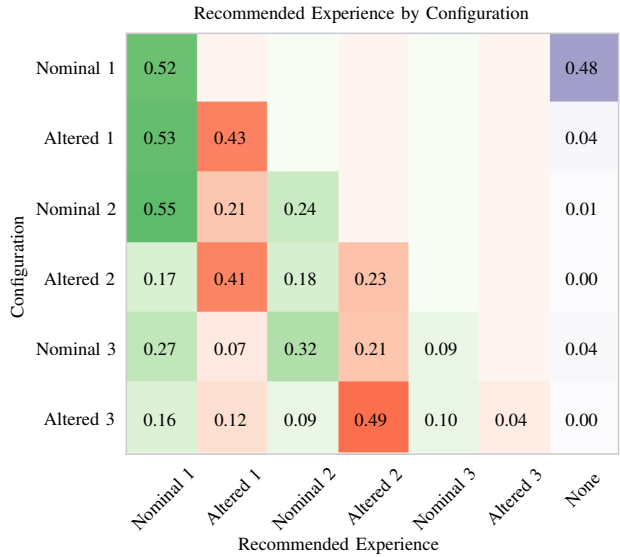
\caption{This figure shows the distribution of recommended experiences by configuration when the vehicle was in each configuration. Each row corresponds to three runs conducted in the same configuration. Each column shows the proportion of experiences recommended from each previous set of runs. The color of the column indicates the vehicle configuration for those runs. Green is for the \textbf{{\color{OliveGreen}\textit{nominal}}} configuration, red is for the \textbf{{\color{red}\textit{altered}}} configuration, and purple is for when \textbf{{\color{Purple}\textit{none}}} of the previous runs were recommended.}
\label{fig:ExpRecConfMtx}
\end{figure}

\begin{figure}
\centering
\scriptsize
\def\svgwidth{0.475\textwidth}
\graphicspath{{figs/Feb24_results/}}
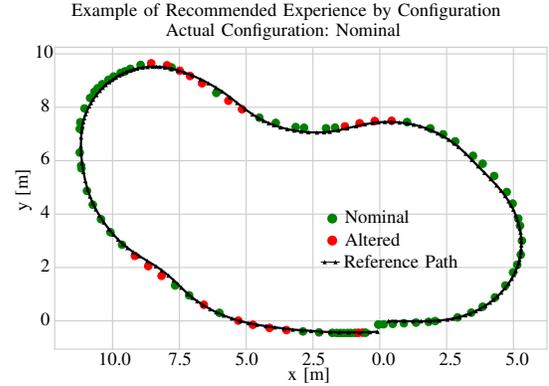
\caption{A top-down view of the path showing the experiences recommended by configuration as the vehicle moves along the path while the vehicle is in the \gtext{\textbf{\textit{nominal}}} configuration. Red and green markers indicate when experiences were recommended from the \rtext{\textbf{\textit{altered}}} and \gtext{\textbf{\textit{nominal}}} configurations, respectively. The reference path is shown in black.  This shows that experiences are primarily recommended from a different configuration along straight sections of the path which is when the dynamics in the two configurations are the most similar.}
\label{fig:ExpRecTopDown}
\end{figure}

Figure \ref{fig:ExpRecConfMtx} shows that the proposed method prefers experiences from the same configuration as the vehicle's current configuration and relies heavily on experiences several runs in the past. This demonstrates that it is beneficial to store not only the most recent experiences, but also many experiences from the past in order to leverage experiences from the same configuration. 

Some of the time, experiences are chosen from runs with a different configuration. Figure \ref{fig:ExpRecTopDown} shows that this is primarily along straight sections of the path where the vehicle is not turning and therefore the dynamics are similar. On all sections of the path with sharp corners where the dynamics are the most different between configurations, the proposed method prefers experiences from the same configuration.

The proportion of time that the proposed algorithm rejects all previous runs decreases rapidly over time. The highest proportion is during the first three runs (Nominal 1 in Fig. \ref{fig:ExpRecConfMtx}) where during the first run, there are no previous runs to draw experiences from, and during the second and third run, the algorithm rejects all runs 17\% and 11\% of the time, respectively. After this initial phase of adaptation, the algorithm manages to find relevant experience over 89\% of the time for each run.

\subsection{Closed  Loop Performance, Long Term Experiment}

Our second experiment was to demonstrate the benefit of the proposed method during long term operation with frequent changes in the operating conditions. The configuration was switched every two runs giving the baseline method one opportunity to adapt before changing the configuration again. 

\begin{figure}
\centering
\scriptsize
\def\svgwidth{0.475\textwidth}
\graphicspath{{figs/Feb27_results/}}
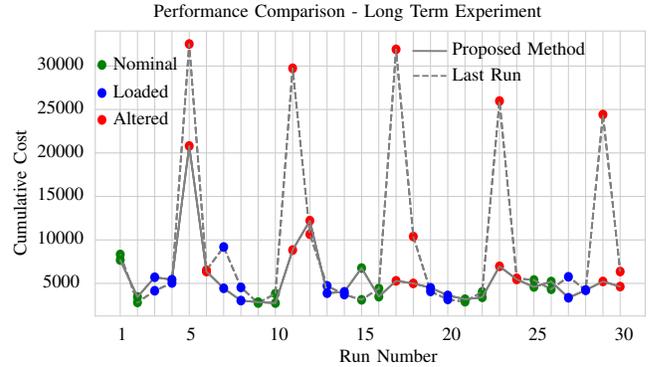
\caption{This figure compares the sum of the control cost for the actual states/inputs over 30 runs in varying configurations when using the proposed method compared to the baseline method. Colored markers indicate the configuration for each run. The vehicle cycled between the \gtext{\textbf{\textit{nominal}}} configuration (green), the \btext{\textbf{\textit{loaded}}} configuration (blue), and the \rtext{\textbf{\textit{altered}}} configuration (red).}
\label{fig:ExpRecLongTerm}
\vspace{-0.25cm}
\end{figure}

Figure \ref{fig:ExpRecLongTerm} shows the cumulative control cost over thirty runs of a similar course to the first experiment. On average, the proposed method results in a 37\% lower cumulative control cost compared to the baseline method primarily due to transitions to the \textit{altered} configuration, which is the most distinct of the three. 

In addition, the number of times the algorithm rejected all candidate runs decreased dramatically after the initial adaptation just like in the first experiment. For the first three pairs of runs, all runs were rejected 55\%, 19\% and 4\% of the time, respectively. After this, the algorithm found matching experiences 97\% of the time with the exception of the second pair of runs in the altered configuration, where it rejected all runs 9\% of the time.

\section{Discussion}

Our method requires that the GP is a good model for the dynamics in each configuration to work well. We conducted experiments in more challenging off-road environments and the GP did not achieve lower M-RMSE than the prior even when learning only from runs in the same configuration. In these cases, our approach could not improve performance simply by changing the experiences in the model. It is important to note that during these runs, the vehicle remained within path tracking constraints so the model error did not jeopardize the safety of the vehicle, it just did not improve the performance. Improving the underlying controller in this way is beyond the scope of this paper, however our method is applicable to any model-based controller that relies on local probabilistic models of the dynamics. 

In addition, our method requires that the dynamics over the previous section of the path are a good indicator of the dynamics on the upcoming section of the path. The operating conditions are always sampled discretely (by run) but can be continuous (e.g. the factor we multiply turn rate commands). Although the method infers by run, which is discrete, it will still work for continuous variables that vary consistently between runs. 

Our method also requires large changes in the dynamics to produce a noticeable improvement. The dynamics in the \textit{loaded} and \textit{nominal} configurations were actually quite similar, so using data from one or the other did not produce large enough changes to be noticeable over the course of a run. If a new configuration with half the load was added, our method would automatically determine that it was similar to the loaded or nominal configurations and leverage data from runs in those configurations.

%
\section{Conclusion}

In this paper, we presented a new, principled method for experience recommendation for long term, GP-based, safe learning control. We demonstrated in closed loop experiments how this method can be used to improve the performance of a controller conducting repeated traverses of a path when the dynamics switch between distinct configurations. This enables the controller to maintain high performance when re-visiting operating conditions that have been seen before and safely learn new dynamics when new operating conditions are encountered. 

In future work, we aim to address the case where the dynamics along the previous section of the path match two or more runs with different dynamics on the upcoming section of the path. The current approach would take a moment to disambiguate the two during which the prediction performance would degrade. Ideally, we would like to avoid this and pass on this information to the controller.

\addtolength{\textheight}{-12cm}   

\bibliographystyle{unsrt}
\bibliography{bib/mm_learning.bib}

\end{document}